\documentclass{article}

\usepackage[nonatbib, preprint]{neurips_2020}

\usepackage[utf8]{inputenc} 
\usepackage[T1]{fontenc}    
\usepackage{hyperref}       
\usepackage{url}            
\usepackage{booktabs}       
\usepackage{amsfonts}       
\usepackage{nicefrac}       
\usepackage{microtype}      
\usepackage{amsmath,amssymb,amsthm}
\usepackage{graphicx}
\usepackage{subcaption}
\usepackage{xcolor}
\usepackage{algorithm}
\usepackage{algpseudocode}

\usepackage[square,sort,comma,numbers]{natbib}

\title{Balancing Robustness and Fairness via Partial Invariance}

\author{%
  Moulik Choraria\thanks{Corresponding author: \texttt{moulikc2@illinois.edu}} \\
  University of Illinois at Urbana-Champaign\\
  \And
  Ibtihal Ferwana \\
  University of Illinois at Urbana-Champaign\\
  \And
  Ankur Mani \\
  University of Minnesota, Twin Cities\\
  \And
  Lav R. Varshney \\
  University of Illinois at Urbana-Champaign\\
}

\begin{document}

\maketitle

\begin{abstract}
The Invariant Risk Minimization (IRM) framework aims to learn invariant features from a set of environments for solving the out-of-distribution (OOD) generalization problem. The underlying assumption is that the causal components of the data generating distributions remain constant across the environments or alternately, the data ``overlaps'' across environments to find meaningful invariant features. Consequently, when the ``overlap'' assumption does not hold, the set of truly invariant features may not be sufficient for optimal prediction performance. Such  cases  arise  naturally  in  networked settings  and  hierarchical  data-generating models, wherein the IRM performance becomes suboptimal. To mitigate this failure case, we argue for a \emph{partial invariance} framework. The key idea is to introduce flexibility into the IRM framework by partitioning the environments based on hierarchical differences, while enforcing invariance locally within the partitions. We motivate this framework in classification settings with causal distribution shifts across environments. Our results show the capability of the partial invariant risk minimization to alleviate the trade-off between fairness and risk in certain settings. 
\end{abstract}

\section{Introduction}
In standard machine learning frameworks, models can be expected to generalize well to unseen data if the data follows the same distribution \cite{vapnik2013}. However, it has been noted that distribution shifts during test time (due to data being from different sources, locations or times) can severely degrade model performance \cite{lakeRTGS2017, marcus2018}.  For instance, in a computer vision task to classify camels and cows, the authors (\cite{beeryGP2018}) showed that during testing, the model with perfect training loss misclassified cows as camels during test time, if the image background was that of a desert. The issue arose due to the model picking up a strong but spurious correlation: during training, most cow images had green pastures, whereas camel images had deserts. Consequently, the model began to associate cows to green pastures and camels to deserts, leading to misclassification during test time. The key observation is that the vanilla classifiers are often prone to exploiting such spurious features during training (in this case, the image background), which can then hamper its performance in out-of-distribution (OOD) tasks. To fix this problem, multiple lines of research have explored the idea of enforcing invariance during learning \cite{arjovskyBGP2020, gulrajaniL2020, ahuja_erm_irm2020, kruegerCJ2021}, to find invariant, causal, features that are independent of the input distribution which may vary across training environments \cite{scholkop2012}. With respect to the vision task considered before, the features specific to the cow in the image (color, shape, etc.) would correspond to invariant features. Therefore, we can obtain a robust predictor by capturing only invariant features, which inevitably capture some causal relationships about the data and thus generalize well in OOD tasks.

The Invariant Risk Minimization (IRM) framework is a promising step to address OOD generalization \cite{arjovskyBGP2020}, which consists of learning a data representation $\phi$ that captures invariant features, followed by training a classifier on the invariant features such that it is close to optimal for all environments. The IRM classifier can perform well in OOD tasks if the conditional expectation of the target given the representation features, does not vary across environments  \cite{arjovskyBGP2020, ahuja_erm_irm2020}, which intuitively means that the data distributions overlap across environments. Due to the additional constraint that the IRM classifier should be optimal for all environments, there is an implicit notion of group fairness in IRM that encourages similar performance across environments, which has been explicitly studied in follow-up works \cite{kruegerCJ2021,RobertCMZ2020}. However, it is unclear how IRM performance is affected if environments do not  overlap adequately. Nevertheless, it is essential to overcome any potential shortcomings of IRM in this setting, since minimally overlapping environments often occur in real-world applications. For instance, in discussions in ethnic community forums, while some linguistic features are shared among communities, there are additional features that are specific only to certain intra-community comments but not outside the community \cite{gallacher2021,ManiV2021}. Thus, one would expect the performance in a downstream task such as hate-speech detection would degrade if such intra-community (non-invariant) features are discarded by the IRM  classifier.

In this work, we propose a \emph{partial invariance} framework that creates partitions within environments, thus introducing the flexibility within IRM to learn non-invariant features locally within the partitions. We show that the partially invariant solution can help improve risk-fairness trade-offs in different distribution shift settings. The organization of this paper is as follows: we describe the related literature in section \ref{sec:rw}, and in section \ref{sec:model}, we explain the details pertaining to the proposed partial invariance framework . In section \ref{sec:res}, we conduct experiments to evaluate the efficacy of our framework, and we provide some the subsequent discussion in section \ref{sec:discussion}. Finally, we highlight possible applications and future work in section \ref{sec:future}. 

\section{Related Work}
\label{sec:rw}
Many recent approaches aim to learn deep invariant feature representations: some focus on domain adaptation by finding a representation whose distribution is invariant across source and target
distributions \cite{shaiBCKPV2010, kunGS2015}, while others focus on conditional domain-invariance  \cite{gongZLTGS2016, liGTLT2018}. However, there is evidence that domain adaption approaches are insufficient when the test distribution is unknown and may lie outside the convex hull of training distributions \cite{leeR2018,duchiGN2018, mohriSS2019}.

The focus of this work is the Invariant Risk Minimization (IRM) framework, proposed in \cite{arjovskyBGP2020}. The framework relates to domain generalization where access to the test distribution is not assumed. It is rooted in the theory of causality \cite{scholkop2012} and  links to invariance, which is useful for generalization \cite{petersBM2016,heinzeCM2018}. In \cite{ahujaSVD2020}, the authors reformulate IRM via a game-theoretic approach, wherein the invariant representation corresponds to the Nash equilibrium of a game. While the IRM framework assumes only the invariance of the conditional expectation of the label given the representation, some follow-up works rely on stronger invariance assumptions \cite{xieYCLSL2021, mahajanTS2021}.

Several follow-up works comment on the performance of the invariant risk framework, and compare against standard Empirical Risk Minimization (ERM). It has been noted that carefully tuned ERM outperforms state-of-the-art domain generalization approaches, including IRM, across multiple benchmarks \cite{gulrajaniL2020}. The failure of IRM may stem from the gap between the proposed framework and its practical ``linear'' version (IRMv1), which fails to capture natural invariances \cite{KamathTDS2021}. The authors of \cite{rosenfeldRP2020} demonstrate that a near-optimal solution to the IRMv1 objective, which nearly matches IRM on training environments, does no better than ERM on environments that differ significantly from the training distribution. The authors of \cite{ahuja_erm_irm2020} argue that IRM has an advantage over ERM only when the support of the different environment distributions have a significant ``overlap''.  However, unlike previous work which aim to minimize OOD risk, our primary focus is to understand the trade-off between risk minimization and fairness across environments in the setting when such a significant ``overlap'' may not exist. Consequently, we analyze how IRM behaves in these settings and explore alternatives via the notion of partial invariance. The notion of fairness considered in this paper relates to \emph{group sufficiency} \cite{liuSH2019}, which is intricately linked with the IRM objective. In fact, several works build upon the IRM framework by explicitly minimizing a quantity resembling this notion of fairness for achieving invariance \cite{kruegerCJ2021, creagerJZ2021}. 

\section{The Partial Invariance Framework}
\label{sec:model}

In this section, we describe a simple experiment using Gaussian mixture models (GMMs) to clearly highlight some of the pitfalls of IRM that are encountered in more complicated practical settings. For completeness, we first state the IRM formulation. 

\paragraph{IRM Framework} 
Consider a set of training environments $e \in \mathcal{E}_{tr}$ and datasets $\{X^e_i, Y^e_i\}_{i=1}^{n_e}$ sampled in each environment from distribution $P_e(X_e, Y_e)$, with number of samples $n_e$, with $X^e_i \in \mathcal{X}$, $Y^e_i \in \mathcal{Y}$ and consider a predictor function $f: \mathcal{X} \rightarrow \mathcal{Y}$. The IRM framework aims to find a predictor $f$ which generalizes well for unseen environments. The goal is to find a data representation $\Phi :  \mathcal{X} \rightarrow \mathcal{Z}$ such that the predictor $w: \mathcal{Z} \rightarrow \mathcal{Y}$ on top of $\Phi$ is invariant across all environments $\mathcal{E}_{tr}$.\footnote{Formally, for all $z \in \mathcal{Z}$ that lies in supports of both $\Phi(X^e), \Phi(X^{e'})$, we want that $E_{P_e}[Y^e |\Phi(X^e) = z] = E_{P_{e'}}[Y^{e'} |\Phi(X^{e'}) = z],\ \forall\ (e, e') \in \mathcal{E}_{tr}$. Such an invariant $\Phi$ allows for the existence of a predictor $w$ which is simultaneously optimal for all environments, and this $w$ is the one that best approximates the conditional expectation of the label.} To achieve this, the training objective for IRM can be summarized as:
\begin{align*}
&\min_{\Phi, w} \sum_{e\in \mathcal{E}_{tr}} \mathcal{R}_e(w \circ \Phi) \tag{IRM}\\ 
\mbox{s.t.} & \ w \in \text{arg}\min_{\tilde{w}} {R}_e(\tilde{w} \circ \Phi)\ \forall\ \tilde{w} \mbox{,} 
\end{align*}

where $R_e(f) = E_{X_e, Y_e}[l(f(X_e), Y_e)]$ represents the risk in environment $e$, pertaining to a loss function $l$ like the mean-squared or the cross-entropy loss. To simplify the bi-level optimization, the authors of \cite{arjovskyBGP2020} propose the following simplification (IRMv1):
$$\min_{\Phi, w} \sum_{e\in \mathcal{E}_{tr}} \mathcal{R}_e(w \circ \Phi) +  \lambda||\nabla_w \mathcal{R}_e(w \circ \Phi) ||^2$$
with regularizer $\lambda$.
In the subsequent discussion, we will describe our experimental setup. As mentioned previously, the pitfalls of IRM, including the shortcomings of (IRMv1) in finding predictors with better OOD risks have been studied before \cite{KamathTDS2021, rosenfeldRP2020}. We take an alternate approach where we highlight that even if IRM (IRMv1) works as intended, the underlying data-generating distribution may not allow finding a `good' solution, from both classification risk and fairness considerations. Simultaneously, we will demonstrate how the partial invariance approach can help in this regard.

\begin{figure}[h]
    \centering
    \includegraphics[width=0.7\linewidth]{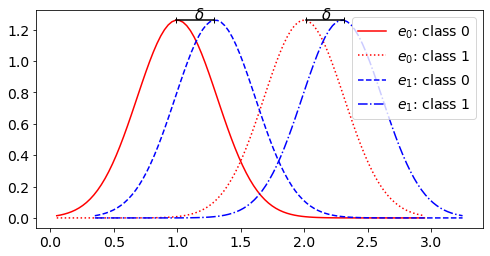}/
\caption{Illustration of the Gaussian mixture distributions for $N=2$, highlighting the role of cross-environmental overlap parameter $\delta$. Here, $(\mu_1, \mu2) = (1, 2)$,  $\sigma=1$, and $\delta=0.3$.}
\label{fig:gaussians_on_line}
\end{figure}

\paragraph{Problem Setup} 
Consider a binary classification task on univariate Gaussian mixtures on an interval of $\mathbb{R}$. Each GMM is parametrized as $g(\mu_1, \mu_2, \sigma)$, i.e.\ two Gaussians with class means $(\mu_1, \mu_2)$ and equal variance $\sigma$. In our setup, we consider $N$ environments wherein each environment in  $\{e_i\}_{i=0}^{N-1}$ is characterized by exactly one Gaussian mixture as $g(\mu_1+i\delta, \mu_2+i\delta, \sigma)$. Therefore, $e_0$ corresponds to $g(\mu_1, \mu_2, \sigma)$ and $e_N$ corresponds to $g(\mu_1+(N-1)\delta, \mu_2+(N-1)\delta, \sigma)$, see Fig.~\ref{fig:gaussians_on_line} for an illustration. Then $\delta$, along with $\sigma$, controls how \textit{far} apart the individual environments are, i.e. the \textbf{\textit{cross-environmental overlap}}. For each environment, we consider a threshold-based classifier. The loss is the (0-1) classification loss, which yields a risk as a continuous function of the threshold $t$. Intuitively, for $N=2$, if $\delta$ is very large, any predictor that does well on the first environment will incur the a very high classification risk in the second environment. 

\paragraph{Motivation}

One would argue that such a problem setting is naturally unsuitable for studying the nature of the IRM solution. Indeed with a single IRM predictor, it follows that with $\Phi(x) = x$, any predictor (threshold) cannot be simultaneously optimal for all $N$ environments, unless $|t| \rightarrow \infty$ and thus violates the assumption on the existence of a `meaningful' IRM solution. However, we argue such situations can naturally arise in standard high-dimensional learning tasks, where 
existence of an IRM solution that is both fair and meaningful may not be guaranteed a priori. Hence the learned $\Phi$, either due to the nature of the data distribution or lack of representation power or simply a failure of the learning algorithm, may yield conditional distributions across environments that show little overlap. Therefore, it is important to analyze and develop countermeasures against this failure case. 

\paragraph{Partial Invariance}
It is clear from the previous discussion that in certain settings, IRM may not work as intended. Looking at the GMM problem in particular, the issue seems to arise from forcing the invariant predictor to be optimal in non-overlapping environments. One intuitive proposal is to partition the environments such that the environments within a particular partition are sufficiently ``close''. Consequently, we can now enforce invariance within each partition. And that is exactly the idea of partial invariance. For now, we stick to the GMM settings as described above and we enforce partial invariance via optimizing the IRMv1 objective, within each partition. We  roughly characterize the Partial-IRM approach below. 

\begin{algorithm}
\caption{Partial-IRM algorithm (rough)}\label{alg:cap}
\begin{algorithmic}
\Require partitions $\{p_i\}_{i=1}^k$, $p_i \subseteq \mathcal{E}_{tr}$, $\sum_{i} |p_i| = |\mathcal{E}_{tr}|$, $p_i \cap p_j = \phi\ \forall\ \ (i,j)$
\For{$\{p_i\}_{i=1}^k$}\\
    $w_i, \Phi_i \leftarrow \text{IRM}(p_i)$
  \Comment{Impose IRM objective over $e$ $\in$ $p_i$, obtain $(w_i, \Phi_i)$ for partition $p_i$}
\EndFor
\end{algorithmic}
\end{algorithm}

It follows from the prior discussion that partial invariance will be  most effective when the environments within each partition are sufficiently close in terms of the distribution, and therefore may be inferred from the environments by minimizing some suitable divergence criterion. For our Gaussian mixtures however, fixing the size of each partition makes the environment assignment intuitive i.e.\ we can simply group together consecutive environments. In later sections, we will discuss how this formulation could fit in the context of causality and invariant features.

\paragraph{Performance Metrics}

As mentioned previously, our focus is not on IRM and learning invariant features for improving OOD performance. In fact, within our GMM setup, we assume access to all environments of interest (i.e. $\{e_i\}_{i=0}^{(N-1)}$). Consequently, our performance metrics depend only on the different $e_i$'s. Within this framework, we analyze and compare the IRM and Partial-IRM solutions on a) classification risk and b) fairness. Given a threshold $t$, we assume a uniform measure on the set of environments and thus, the classification risk may be computed as follows:
\begin{equation}
    R_g(t) = \frac{1}{N} \sum_{i=0}^{(N-1)} R_{e_i}(t) \mbox{.}
    \label{eq:gen_risk}
\end{equation}
Here $R_{e_i}(t)$ denotes the classification risk due to the choice of threshold $t$ for environment $e_i$ (i.e. the risk on the GMM parametrized as g($\mu_1+i.\delta, \mu_2+i.\delta, \sigma$)). Next, we estimate fairness based on the notion of "Group-Sufficiency" --satisfied when
$E[y|S(x), e] = E[y|S(x), e_0]$  for all $e, e_0$, where $e$ denotes membership of some sensitive class and $S(x)$ represents a score metric. It roughly equates to equalizing the risks across environments. For a choice of threshold $t$, we measure fairness using the V-REx penalty\cite{kruegerCJ2021}, which roughly estimates the variance of the risks across environments (lower variance implies higher fairness): 
\begin{equation}
    F_{V}(t) = \frac{1}{2N^2} \sum_{i=0}^{(N-1)} \sum_{j=0}^{(N-1)}(R_{e_i}(t)-R_{e_j(t)})^2 \mbox{.}
    \label{eq:fairness}
\end{equation}


\section{Experiments and Results}
\label{sec:res}
Our goal is to test the effect of creating partitions on the performance of invariant predictors, characterized by both risk and fairness. Given 24 environments, we test with different number of partitions $p \in \{1,2,3,4,6,12,24\}$, with equal sized partition for simplicity. The IRM case corresponds to $p=1$ with a single classifier for all environments, whereas the ERM case is $p=24$ with a classifier for each environment. Based on the Gaussian mixture model, with $\mu_1 = 1$ and $\mu_2=2$, we experiment with four combinations of intra-environment separation $\sigma$ (0.1 or 1), and cross-environments overlap $\delta$ (0.1 or 1), noting that the highest overlap between environments is achieved with large $\sigma$ ($\sigma$ = 1) and a small $\delta$ ($\delta$ = 0.1). We set the regularization $\lambda$ to be 10, which for our problem suffices for IRMv1 to mirror the true IRM objective. We learn threshold classifiers, by minimizing the IRMv1 objective for the environments within that partition. In the case of partition size being equal to 1, we consider the ERM objective by setting $\lambda = 0$.

\begin{figure}[!ht]
\centering
    \begin{subfigure}{.45\textwidth}
        \includegraphics[width=\linewidth]{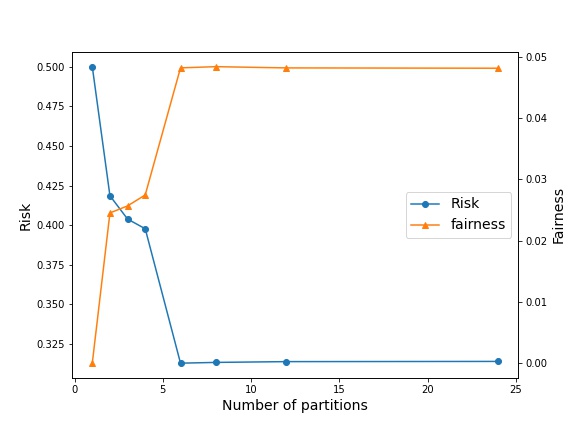}
        \caption{Large separation, high overlap}
        \label{fig:case1_5}
    \end{subfigure}
    \begin{subfigure}{.45\textwidth}
        \includegraphics[width=\linewidth]{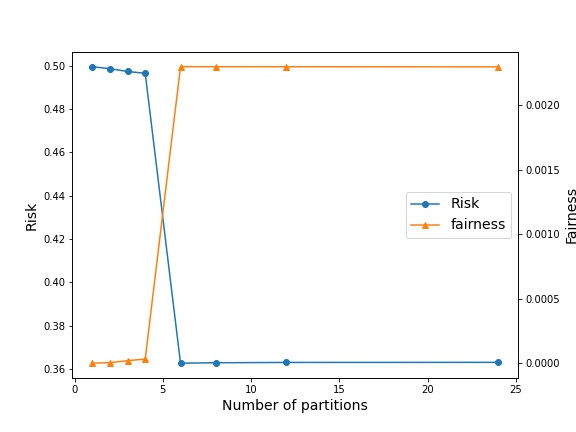}
        \caption{Minimal separation, high overlap}
        \label{fig:case2_5}
    \end{subfigure}
  \begin{subfigure}{.45\textwidth}
        \includegraphics[width=\linewidth]{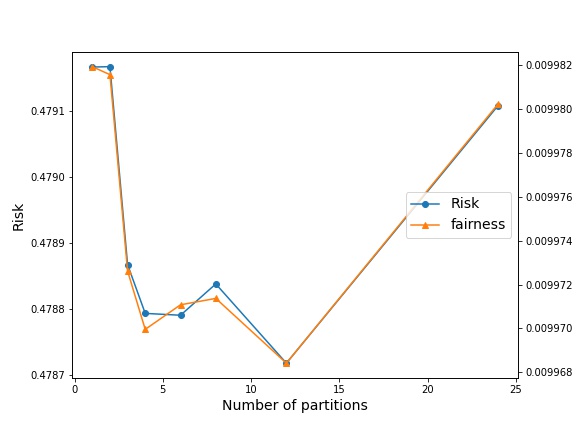}
        \caption{Large separation, low overlap}
        \label{fig:case3_5}
    \end{subfigure}
    \begin{subfigure}{.45\textwidth}
        \includegraphics[width=\linewidth]{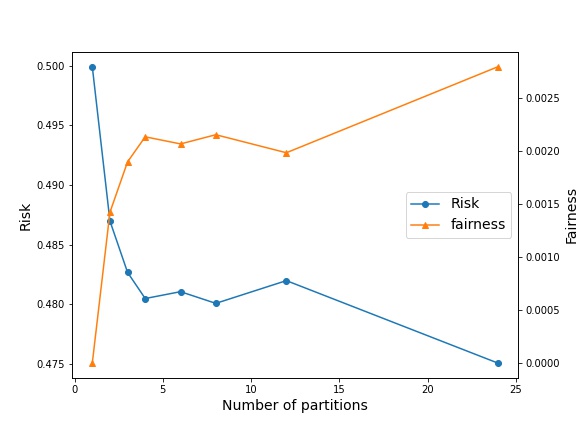}
        \caption{Minimal separation, low overlap}
        \label{fig:case4_5}
    \end{subfigure}
    \caption{Experiments with Gaussian mixtures for different $\delta, \sigma$ values. Lower values of risk and fairness variance indicate better performance and fairness respectively.}
    \label{fig:lambda_5}
\end{figure}

\paragraph{High overlap (small $\delta)$} A small value of $\delta$ means that environments show a stronger degree of overlap. In this setting, we observe that the IRM solution prioritizes fairness over risk. Herein, partial invariance allows us to trade-off performance for reduced fairness. However, in certain settings wherein the classes in the environment are not well separated (large $\sigma$), one can improve significantly improve the performance by paying a much lower relative cost in terms of fairness (Fig.~\ref{fig:case2_5}). This also demonstrates that P-IRM introduces flexibility in terms of balancing risk and fairness.    


\paragraph{Low Overlap (large $\delta)$} Next, we consider a larger value of $\delta$, meaning a low environmental overlap. In both cases, we observe that the degree of change in the risk/fairness values is lower than the previous case. However, we point out the curious case of well separated environments with low overlap, wherein P-IRM can seemingly improve both risk and fairness simultaneously (Fig.~\ref{fig:case3_5}). Since we expect such settings to arise in real world problems, our results offer us some intuition towards how P-IRM can improve performance in such settings.

\subsection{The Fairness-Robustness Trade-off}

The previous experiments give us insights into the situations in which partial invariance can be beneficial in controlling the fairness-risk trade-off. Specifically, in settings with a smaller $\delta$ and consequently more overlap, the fairness-risk values change significantly  as one increases the number of partitions. However, one could argue that such a trade-off could be achieved by simply varying the $\lambda$ for the IRM solution. But we claim that introducing the notion of partial invariance allows us easier access to a richer solution set on this trade-off, while still retaining desirable properties.

\begin{figure}[h]
\centering
\begin{subfigure}{.49\textwidth}
  \centering
  \includegraphics[width=.98\linewidth]{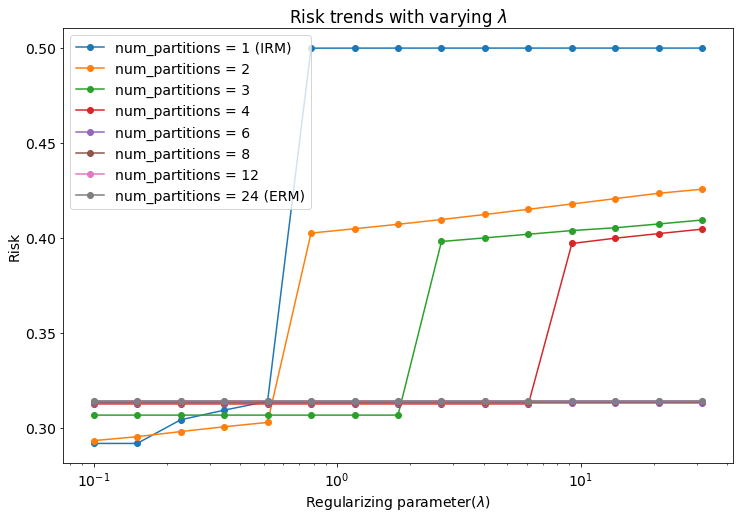}
  \caption{Risk of different partitions} 
  \label{fig:lambda_risks}
\end{subfigure}%
\begin{subfigure}{.49\textwidth}
  \centering
  \includegraphics[width=.98\linewidth]{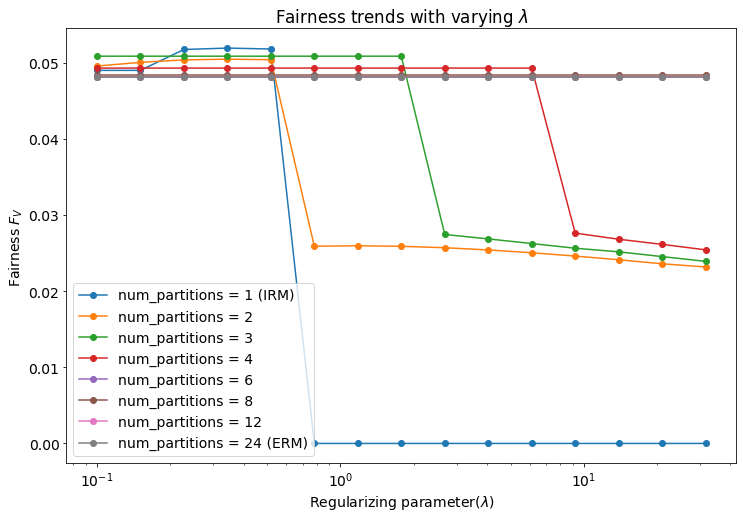}
  \caption{Fairness of different partitions} 
  \label{fig:lambda_fairs}
\end{subfigure}
\caption{Effects of adjusting the regularizing parameter $\lambda$ on risk and fairness of IRM, P-IRM, and ERM solutions.}
\label{fig:trade_off}
\end{figure}

To show this, we consider the high overlap setting in which the individual Gaussians within an environment are easily separable, i.e. $\delta = 0.1 = \sigma$. The rest of the setup is exactly the same as the previous experiment. But here, we fix our data generating distribution and instead vary $\lambda$ on the log scale. Then, for each value of $\lambda$ and given partition size, we measure the corresponding risk and fairness values for the obtained optimal threshold. The obtained plots are presented in Fig.\ref{fig:trade_off}. 

We notice that in the regions corresponding to sufficiently high values of $\lambda$, the Partial-IRM solutions allow us to operate on different curves corresponding to risk-fairness trade-offs, remarkable similar to receiver operating characteristic (ROC) curves in classical decision theory. Additionally, we can switch between different levels of risk/fairness by switching between the number of partitions. Note our predefined constraints on equal-sized partitions restricts the choice of partition size to integer factors of $N$. We speculate that we might achieve even more flexibility in certain other settings, such as when both the environments and partitions being defined as continuous intervals. And the key observation here is that while IRM may be able to find similar trade-offs by reducing $\lambda$ (the sharp slope region for the blue curve, Fig.~\ref{fig:lambda_risks}), it does so by \textit{compromising} on the basic essence of invariance. In the discussion, we will elaborate on how this might not be desirable in certain settings.

\section{Discussion}
\label{sec:discussion}

The experiments in the previous section offered us some interesting insights. We saw that when the cross-environments overlap is low, Partial-IRM offers only marginal benefits over standard IRM and ERM. This is expected because a lower similarity between environments implies that there is less common information that may be exploited. Additionally, there may exist some interesting cases wherein we can simultaneously improve upon on both fairness and classification risk, by adopting the partial invariance approach. This is somewhat surprising and certainly merits additional investigation. Next, we considered the high overlap settings. Here, the role of partial invariance in navigating the risk-fairness trade-off comes to fore. We elaborate upon this idea in the subsequent experiment, wherein we observe that by changing the size of the partition, we can switch between different levels of the risk-fairness trade-off. To achieve the same flexibility within the IRM framework, we need to operate within regions where the IRMv1 solution shows large deviations from the original IRM formulation (by reducing $\lambda$). We believe this can have additional detrimental effects in situations where we would implicitly need invariance. We demonstrate one such setting below. 

\subsection{The Case for Partial IRM}
In section ~$\ref{sec:res}$, we observed that the notion of partial invariance allowed us to explore the inherent trade-off between risk minimization and fairness and in certain cases, simultaneously improve upon both risk and fairness as compared to the IRM solution. However, it is not yet clear as to how partial invariance fits in the ``causal'' picture of invariant risk minimization. We attempt to shed some light on this by slightly modifying the motivating regression task in the original IRM formulation \cite{arjovskyBGP2020}.

Consider the following data generation model, where $\sigma(e) \in [0, \sigma^2_{max}]$. The target $Y$ is to be regressed based on observed values of $X = (X_1, X_2, X_3)$. We consider two independent environment variables $(E, e)$ at play, where $e$ affects the individual distributions of the set of observables $X$, whereas $E$ controls the marginal distribution of $Y$ given $X_2$ via the random variable $P$:
\begin{equation*}
    \begin{gathered}
     X_1 \leftarrow N(0, \sigma(e)^2), X_2 \leftarrow N(0, \sigma(e)^2)\\
     E \leftarrow \textbf{U}(\{1, 2, 3\}), P(E) = f(E)\\
     Y \leftarrow X_1 + P(E).X_2 + N(0, \sigma(e)^2)\\
     X_3 \leftarrow Y + N(0, 1)
    \end{gathered}
\end{equation*}
We consider the prediction model to estimate $Y$ as $\hat{Y} = \alpha_1 X_1 + \alpha_2 X_2 + \alpha_3 X_3$. Then. it follows that if $\alpha_3 > 0$, the regression coefficients will explicitly depend on $\sigma(e)$ and the OOD risk of the model can grow without bound, as noted in \cite{arjovskyBGP2020}. Within the IRM framework, the only feasible representation $\Phi$ that yields invariant predictors across $(E, e)$ is $\Phi(X) = X_1$ and the corresponding regression coefficients are $(1, 0, 0)$ respectively. While this solution works and generalizes for OOD tasks w.r.t.\ $e$, we lose in terms of performance by discarding $X_2$. In addition, deviating from the IRM solution via IRMv1 (as in the previous section) is not feasible since we require invariance w.r.t $e$ for a finite risk in worst-case environments.

Now note that the notion of partial invariance offers an intuitive solution; We could first create partitions across $E$ and then apply IRM to obtain invariance w.r.t e, potentially allowing us to get the best of both worlds. However, as in the case with GMMs, the idea implicitly assumes that the P-IRM framework can indeed partition across $E$. In general, it is quite hard to even achieve this distinction between causal and non-causal features, without prior information on the data-generating distributions. Fortunately, it seems that it is possible to surmise this information under certain settings that arise naturally and a study of precisely these settings will be the focus of our future work.


\section{Future Work/Applications}
\label{sec:future}
Distribution shifts naturally arise in real-world settings, particularly when environments with low overlap interact so that even though sub-groups share some common features, they can differ substantially in some other locale-specific features \cite{ManiV2021}. In such settings, the partial invariance framework can better tackle the OOD problem. Similar notions can also be found in recent literature pertaining to equivariant neural networks, in relaxing global symmetry requirements to local gauge symmetries\cite{cohenWKW2019}.

Next, we describe applications where partial invariance could prove to be useful. In online forums, identity-related discussions in which hatred or humiliation is expressed towards certain out-groups may be considered toxic, but similar discussions in a completely in-group discussion may not \cite{gallacher2021}. Also, in forums that usually welcome diversity, explicitly mentioning identities would be acceptable and not toxic, unlike in forums that routinely make discriminatory remarks against minorities.
Existing systems to detect toxicity in online forums \cite{badjatiyaFMV2017, zhangLuo2019} do not, however, consider the environmental differences between forums and tend to perform poorly in more diverse settings \cite{borkanDSTV2019}.  As such, a model that can distinguish the type/purpose of a forum and then infer toxic mentions from this additional context, can be expected improve performance. 

We note that in the current form however, it is unclear how partial invariance can be applied to real data wherein firstly, we need to decide how to partition environments and secondly, we need to infer the partition membership during test time, so as to use the corresponding classifier for optimal prediction for that sample. Therefore, in future work, we want to develop methods to automatically infer partitions from environments, by introducing appropriate notions of distance between environments. 
\bibliographystyle{plain}
\bibliography{full,conf_full, prim_lib} 

\end{document}